\documentclass[letterpaper, 10 pt, conference]{ieeeconf}
\IEEEoverridecommandlockouts
\usepackage{amsmath}
\usepackage{float}
\usepackage{graphicx}
\usepackage{booktabs}
\usepackage{subcaption}
\usepackage{multirow}
\usepackage{url}
\usepackage[noadjust]{cite}
\let\labelindent\relax
\usepackage{enumitem}
\setlist[itemize]{leftmargin=4.5mm}
\setlist[enumerate]{leftmargin=4.5mm}
\usepackage{cuted}     
\usepackage{capt-of}   
\title{\LARGE \bf Robot Motion as a Keyed Communication Channel}
\author{Benjamin Chang$^{1}$, Michael Amir$^{2}$, Manon Flageat$^{2}$, and Amanda Prorok$^{2}$%
\thanks{$^{1}$Department of Engineering, University of Cambridge, United Kingdom.
{\tt\small btyc3@cam.ac.uk}}%
\thanks{$^{2}$Department of Computer Science and Technology, University of Cambridge, United Kingdom.
{\tt\small \{ma2151, mf873, asp45\}@cam.ac.uk}}%
}

\begin{document}
\maketitle
\thispagestyle{empty}
\pagestyle{empty}

\begin{strip}
\centering
\includegraphics[width=\textwidth]{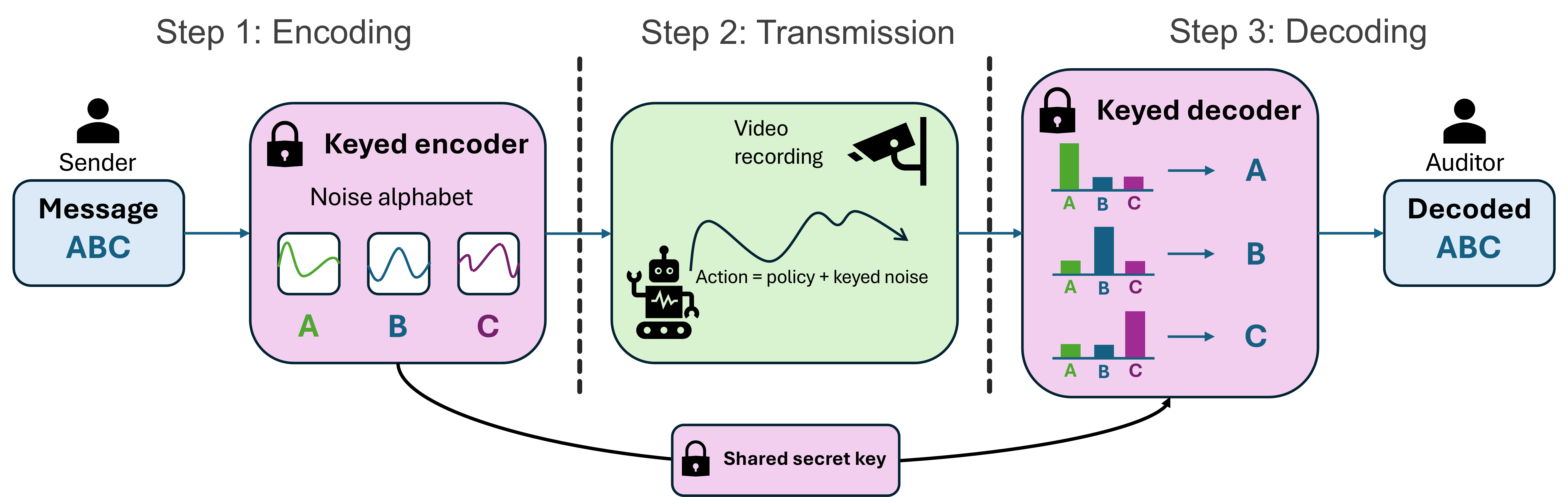}
\captionof{figure}{\textbf{Messaging through motion.} A keyed encoder maps each message symbol to a distinct noise pattern, which is added to a pre-trained policy's actions. An auditor, using the shared secret key, matches observed motion from video against candidate patterns to recover the message.}
\label{fig:intro}
\end{strip}


\begin{abstract} Messages from electronic devices are conventionally received as text, audio, or radio signals. But robots move with rich, articulate motion in the real world, opening up the possibility of transmitting messages through motion itself. In this paper, we consider the problem of \emph{motion-based communication}, where we seek to modify a robot's movements so as to transmit messages detectable from remote sensing (e.g., video or motion capture), without degrading policy performance. We introduce a method for \emph{messaging through motion} capable of encoding arbitrary message content over short payloads---such as an agent's current intent---as noise in any pre-trained policy's actions. This brings a new kind of robustness to robot communication: this `physical' channel complements standard wireless communications channels but does not depend on them, requiring no extra hardware nor the establishment of a direct link to the robot. We systematically characterize the space of encoding schemes and derive design heuristics, then validate them across simulated environments and real-robot deployment; on real robots running at 50 Hz, four robots jointly recover an 8-bit message at an aggregate 0.8 bits/s.
\end{abstract}


\section{Introduction}
This work explores robot communication through motion. Humans already use
motion as a communication medium: body language, gestures, and pantomime all
convey meaning through deliberately legible movements. Robots can communicate in
this way too, and prior work has explored motion that helps human observers
infer a robot's goal or intent~\cite{dragan2013legibility,Dragan-RSS-13}. 
But robots also open a different, less human-like possibility. Because their
motion can be controlled to high precision with small, repeatable stochasticity, robots can use motion not only for visible signaling, but also as a
carrier for keyed messages. We study arbitrary message content, over short
payloads, embedded subtly in
robot motion, such that the behavior looks ordinary and does not disturb task
performance, yet remains recoverable from remote sensor data such as video or
motion capture.

Motion-based transmission becomes increasingly useful as
robots are deployed everywhere. A robot's motion is visible to anyone with a
camera, and phone and CCTV cameras are ubiquitous---so motion is a channel that
reaches everyone, not just operators with dedicated wireless links and
compatible receivers. This opens the message up to bystanders, regulators, and
third-party auditors: an autonomous vehicle could signal which fleet protocol it
is running, or a robot could broadcast its operator ID, all recoverable from an
ordinary video. Motion can also complement conventional communication when wireless channels may be unavailable, jammed, incompatible, or untrusted. In such scenarios, motion can serve as a physical handshake, i.e. an authentication signal binding the channel's identity to a given robot embodiment.

At the same time, not every message should be legible to everyone: privacy
matters, and a robot may need to communicate with a specific auditor without
broadcasting to every camera in range. This calls for a keyed channel, where
only a recipient holding a shared secret key can recover the message. We
introduce \emph{Messaging Through Motion} (MTM): a method that encodes arbitrary, keyed
message content into a robot's motion (Fig.~\ref{fig:intro}), recoverable from remote
observations by a recipient holding the key.\footnote{Project website and code:\\\url{https://sites.google.com/view/motionbasedmessaging}}

Realizing this requires a method that can (i) encode arbitrary message content in reasonable time; (ii) decode reliably from remote sensor data, such as video footage or motion capture traces; (iii) keep the message private, so that decoding without the shared key recovers symbols no better than chance; and (iv) deploy on existing stochastic policies without retraining, architectural changes, or extra hardware.
We present the first method to meet all four requirements, and formalize the underlying design space of this method. Our main insight is that recent advances in remotely detectable policy watermarking, which embeds keyed signals into a robot's behavior that are recoverable from remote sensors, can be extended into a full multi-symbol messaging method. While remotely detectable schemes, such as CoNoCo~\cite{conoco}, recover only a single bit, extending to multi-symbol messages brings new challenges: errors compound across the message, and total transmission time becomes an important constraint.

MTM targets stochastic policies, which map observations to actions, and are common in learning, particularly in reinforcement learning~\cite{sutton1998reinforcement} and learning for robotics. Rather than adding separate stochasticity, MTM shapes the policy's existing noise into a keyed signal at inference time. Prior work~\cite{conoco} shows that, for Gaussian policies, this preserves the policy's marginal action distribution; a guarantee MTM inherits. MTM also applies to deterministic policies by injecting small Gaussian noise, with performance-detectability trade-off controlled by the noise amplitude. We confirm this empirically in Sec.~\ref{app:reward-impact}.

Our contributions are as follows:
\begin{itemize}

    \item \textbf{Motion messaging.} We introduce MTM, a keyed method for embedding arbitrary message content into the action-noise channel of pre-trained stochastic policies at inference time, recoverable from passive remote motion observations (e.g., motion capture).

    \item \textbf{Watermark-to-message construction.} We show how to turn a remotely detectable single-watermark primitive into a multi-symbol communication channel by mapping message symbols to keyed signals and decoding them from video-derived motion traces.

    \item \textbf{Design space.} We formalize the trade-offs between alphabet size, symbol duration, and channel allocation, showing how to choose the fastest scheme that meets a target decoding accuracy.

    \item \textbf{Evaluation.} Across five environments (Discovery, Lunar Lander,
    Reacher, Football, and real-robot RoboMaster~\cite{blumenkamp2024robomaster}),
    the same-message scheme recovers 3-bit payloads in $5.5$--$10.8$~s in
    simulation and a 6-bit team strategy in $52$~s in the adversarial multi-agent
    Football setting; on hardware, the split-message scheme recovers an 8-bit
    message across four robots in $10$~s (Sec.~\ref{sec:results}).
\end{itemize}

\section{Related Work}
\label{sec:related}

\noindent\textbf{Watermarking.}
Digital watermarking embeds hidden signals in media for provenance and ownership, with spread-spectrum methods distributing signals across frequency bands for robustness~\cite{berghel1996protecting,swanson1998multimedia,cox1997secure}; related ideas now appear in generative models~\cite{wen2023treering,kirchenbauer2023watermark,dathathri2024scalable}. In cyber-physical systems, dynamic watermarking injects secret probes into control inputs to detect spoofed sensors~\cite{satchidanandan2016dynamic,ko2016theory}, but assumes access to internal control and privileged internal sensors rather than passive remote observation. Neural-network and policy watermarking similarly target ownership or identification through weights, triggers, safe states, or query behavior~\cite{darvishrouhani2019deepsigns,adi2018turning,behzadan2019sequential,chen2021temporal,szyller2021dawn,huang2025agent}; these approaches do not provide a keyed, remote, multi-symbol communication channel through physical motion.

\noindent\textbf{Remotely detectable robot-policy watermarking.}
Closest to our work, CoNoCo~\cite{conoco} embeds keyed colored noise into a stochastic policy and detects it from video via spectral coherency. CoNoCo establishes the remote watermarking primitive we need, but its decision is binary: whether a known keyed signal is present. MTM builds a communication layer on top of such a primitive, assigning distinct keyed signals to message symbols, and decoding over successive time windows. The naive extension---binary detection time-divided over a single channel---is
our $M{=}2$ single-channel condition, and is never the best scheme at any
operating point we test; reliable messaging requires choosing an alphabet size and channel allocation, the design space we characterize.

\noindent\textbf{Steganography.}
Classical steganography hides messages in a carrier signal, whereas
watermarking usually emphasizes provenance or ownership. A recent line of
work hides messages in the sampling distributions of generative models, e.g.\
diffusion models~\cite{kim2023diffusionstego,jiang2025stegamgm} and
LLMs~\cite{norelli2026calgacus}. These methods embed messages into the
latent noise or token-sampling steps, so that the output remains
statistically indistinguishable from an ordinary sample. MTM extends this
idea from generative models in software to policy execution in the physical
world: the randomness exploited is the exploration noise of a stochastic
policy, and the resulting motion is the carrier observed by a remote auditor.
Unlike these methods, we make no claim of statistical undetectability: our
scheme hides message content from observers without the key, but not the
presence of a watermark.

\noindent\textbf{Legible motion.}
Dragan et al.~\cite{dragan2013legibility,Dragan-RSS-13} shape goal-directed trajectories so
that an observer can quickly infer which goal, from a known set, the robot is
pursuing. By design, the conveyed information is the task goal itself, and it is
expressed by deliberately deviating the trajectory away from the efficient path
so the goal is unmistakable---and is thus readable by anyone watching. Our
scheme instead encodes arbitrary message content, decoupled from the task, as subtle
noise on a channel closed to anyone without the key.

\section{Problem Setting}
\label{sec:problem}
A robot operates in state space $\mathcal{S}$ and action space $\mathcal{A}$, executing a fixed, pre-trained stochastic policy $\pi: \mathcal{S} \times \mathcal{A} \to [0,1]$ on its primary
task. We wish to transmit a message $m$---a sequence of $L$ symbols over an
alphabet of size $M$---to a remote auditor holding a shared secret key $k$, by
adding stochasticity to $\pi$'s actions at inference time. Crucially, the method may not retrain
$\pi$, alter its architecture, or introduce additional hardware. The auditor never sees the
commanded actions; it observes only the robot's resulting motion through remote sensing. The goal is a keyed encoder--decoder pair such that the auditor
recovers $m$ with probability at least $1-\epsilon$, in the shortest
transmission time $T$, while: (i) task performance is preserved, and (ii) decoding with a key other than $k$ recovers $m$ no better than chance.

\section{Preliminaries}
\label{sec:prelim}

Our method builds on Colored Noise Coherency (CoNoCo)~\cite{conoco}, a framework for embedding remotely detectable watermarks into a robot's motion. CoNoCo formalizes the \emph{Physical Observation Gap}: a remote auditor does not observe the
policy's commanded actions, but only their physical consequences (e.g.\ motion in
video). Any watermark must thus survive:
(C1) synchronization uncertainty between the policy's execution clock and the
sensor's sampling rate; (C2) system dynamics filtering commanded actions through
the robot's unknown physics; and (C3) interference from the policy's primary
behavior and from sensor and environmental noise. The policy runs at unknown times $\{T_k\}$ and its executed action
$a_{\mathrm{exec}}(t)$ drives state evolution
$\dot{s}(t) = S_{\mathrm{dyn}}(s(t), a_{\mathrm{exec}}(t))$. A remote sensor
samples at times $\{t_i\}$ (rate $f_g$), yielding the auditor's data: a
\emph{glimpse sequence} 
$\mathcal{G} = (G_i)_{i=0}^{N-1}$ where $G_i = G_{\mathrm{map}}(s(t_i)) + \eta_i$,
and $G_{\mathrm{map}}$ maps state to a remote observation (e.g.\ a velocity
estimate from video) and $\eta_i$ is measurement noise.

CoNoCo~\cite{conoco} addresses (C1)--(C3) by replacing a stochastic policy's exploration
noise with keyed colored Gaussian noise (CGN) concentrated in a secret band
$B$, and detecting it via spectral coherency between the candidate CGN and
$\mathcal{G}$---a metric invariant to LTI dynamics (C2), searched over
candidate policy rates to resolve (C1). CoNoCo's output is binary: a policy
is either identified or not. It is this single-bit case that our method
extends to multi-symbol messages.

We use CoNoCo as a \textit{black-box} remotely detectable watermarking primitive inside MTM. Given a secret key, the primitive generates a statistical signal that can be injected into the stochastic policy's actions. Given a remote sensing-derived motion trace, it returns a score for how strongly that seeded signal appears. MTM turns this single-watermark primitive into a messaging scheme by assigning different seeds to different symbols and decoding each symbol window by choosing the highest-scoring seed. In our experiments, CoNoCo instantiates this primitive using keyed CGN and spectral coherency. MTM only requires these two operations, so another remotely detectable watermarking method could replace CoNoCo should one arise.\footnote{To our knowledge at the time of writing, CoNoCo is the only remotely detectable robot-policy watermark.}
\section{Method}
\label{sec:method}

Our approach, \emph{Messaging through Motion} (MTM), sends a message by introducing remotely detectable, secret key-dependent signals to a robot policy's actions. In stochastic policies, this is done by modifying the pre-existing random noise of the policy rather than adding the noise signal naively~\cite{conoco}, which preserves the policy's marginal action distribution. Each
message symbol is assigned its own noise signal, generated from a secret
key. To send a symbol, the robot introduces the corresponding noise signal to
one or more action channels while running its normal policy.

A remote auditor who knows the key can regenerate the candidate noise
signals. From the observed motion, the auditor checks which candidate
signal is most present, and decodes that as the transmitted symbol. The
message is recovered by repeating this process over successive time
windows. We use CoNoCo~\cite{conoco} as the \textit{watermarking primitive} in all
experiments. It generates the keyed noise signals and scores their presence in
remote observations, using CGN for injection and spectral coherency for
detection. MTM wraps around this primitive by mapping symbols to seeds and combining scores across channels.

MTM has three parts. We first define the available motion
\emph{channels}, i.e. observable degrees of freedom that can carry a
signal (\S\ref{sec:multi-channel}). We then describe the \emph{encoding}:
how message symbols are mapped to keyed noise signals, and how the auditor
decodes them from remote observation (\S\ref{sec:intent-alphabets}). Finally, we define
the \emph{design space}: how to choose the channel strategy, alphabet size,
and time per symbol to transmit a message as quickly and reliably as
possible (\S\ref{sec:design-space}).

\subsection{Channels}
\label{sec:multi-channel}

A \emph{channel} is any observable degree of freedom into which a watermark can be injected, i.e.\ any actuated part that can be tracked from remote sensors. 
For example, channels can correspond to different limbs of a single agent, or to separate agents in a multi-agent setting.
Since different channels have different capacities and reliability, we consider three ways of using them, with $N$ the number of channels:
\begin{itemize}
    \item \textbf{Single-channel}: one channel carries the full message (symbol sequence).
    \item \textbf{Same-message}: all $N$ channels carry the same symbol sequence; detection scores are pooled across channels (averaging or majority vote) to reduce noise.
    \item \textbf{Split-message}: each channel carries an independent part of the message in parallel, increasing time per symbol by a factor of $N$ (total message time unchanged).
\end{itemize}

\subsection{Encoding and Decoding}
\label{sec:intent-alphabets}

\noindent\textbf{Encoding.}
We encode messages over a finite alphabet of size~$M$, where each symbol $i \in \{0, \dots, M{-}1\}$ is associated with a distinct keyed random seed~$k_i$. The watermarking primitive uses $k_i$ to generate the noise signal injected
during that symbol's time window. In our instantiation using CoNoCo as the watermarking primitive, this is a colored Gaussian noise sequence concentrated in a secret
frequency band. Suppose the total time budget available for transmission is $T$. Under $M$-ary encoding, each symbol conveys $\log_2 M$ bits, so larger $M$ requires fewer symbol windows for the same payload, allowing a longer window per symbol. This improves decoding accuracy through noise averaging --- though it also increases confusion risk, as the decoder must distinguish between more candidates.

\noindent\textbf{Decoding.}
The auditor, who shares the secret keys $(k_i)_{i \in \{0, \dots, M{-}1\}}$, decodes each symbol independently. For each candidate symbol~$i$, the auditor computes
the primitive's detection score between the candidate watermark generated
from~$k_i$ and the observed glimpse sequence for that time window, then
selects the highest-scoring one. In our CoNoCo instantiation, this
score is the spectral coherency between the keyed CGN reference and the
observed motion. Repeating this across all symbol windows recovers the full
message.

\subsection{Design Space}
\label{sec:design-space}

Together, \S\ref{sec:intent-alphabets} and \S\ref{sec:multi-channel} define the design axes: channel strategy, alphabet size, and time per symbol.

\noindent\textbf{Design procedure.}
Given a target message accuracy $1-\epsilon$, we select the scheme transmitting
the message in the shortest total time $T$. Let $a_c(t, M)$ be the per-symbol
decoding accuracy on a single channel $c$ at time-per-symbol $t$ and alphabet
size $M$, and $a_\mathcal{I}(t, M)$ the per-symbol accuracy when a subset of
channels $\mathcal{I}$ redundantly carry the same symbol (pooled via averaging
or majority vote); both are measured empirically. Under the working approximation that per-symbol errors are independent, an
$L$-symbol message is decoded correctly with probability equal to
the product of per-symbol accuracies. For each scheme we minimize $T$ subject to
this product meeting the target:
\begin{enumerate}
    \item \textbf{Single-channel:} over $M$ and channel $c$, s.t.\
    $a_c(T/L, M)^L \geq 1-\epsilon$.
    \item \textbf{Same-message:} over $M$ and channel subset $\mathcal{I}$, s.t.\
    $a_\mathcal{I}(T/L, M)^L \geq 1-\epsilon$.
    \item \textbf{Split-message:} over $M$, channel subset $\mathcal{I}$, and
    allocations $\{l_c\}_{c\in\mathcal{I}}$ with $\sum_c l_c = L$, s.t.\
    $\prod_{c\in\mathcal{I}} a_c(T/l_c, M)^{l_c} \geq 1-\epsilon$.
\end{enumerate}
Because the accuracy functions are empirical, each minimization is a direct
search over the measured accuracy table; we select the scheme with the smallest
$T$.

\noindent\textbf{Capacity-based allocation.}
\label{sec:capacity}
For the multi-channel split-message method, it is straightforward to empirically find the
optimum allocation of symbols across the channels for small $L$. However, for large $L$, we propose
a fast heuristic allocation of message symbols across channels using their capacities.
Since we care only whether each symbol is decoded correctly or not, we model each channel as
an $M$-ary symmetric channel with symbol accuracy $a_c$. This model is reasonable because each symbol uses an independent random seed and each window a fresh noise realization, so $a_c$ can be approximated as constant across symbols, and errors as independent across windows. This gives the capacity
of the channel (in bits/symbol):
\begin{equation}
    C_M(a_c) = \log_2 M - H_2(a_c) - (1 - a_c)\log_2(M-1)
    \label{eq:capacity}
\end{equation}
where $H_2(\cdot)$ is the binary entropy function. This correctly vanishes at
$a_c = 1/M$, where the decoder conveys no information about the transmitted
symbol. So if each channel $c$ is allocated $l_c$ message symbols, the maximum
number of reliably decodable message symbols per channel is
\begin{equation}
    R_c = l_c \cdot C_M(a_c),
    \label{eq:Bc}
\end{equation}
i.e.\ more symbols are allocated to more reliable channels. It should be noted that this is an
approximation: the capacity-based allocation assumes a target accuracy of $100\%$, the same total
number of symbols per channel (coded and uncoded), and an asymptotically long message.
The single-channel accuracy curves used
to select allocations are not held out: calibration and evaluation share seeds.


\section{Results}
\label{sec:results}

\begin{figure*}[!t]
\centering
\includegraphics[width=0.96\textwidth]{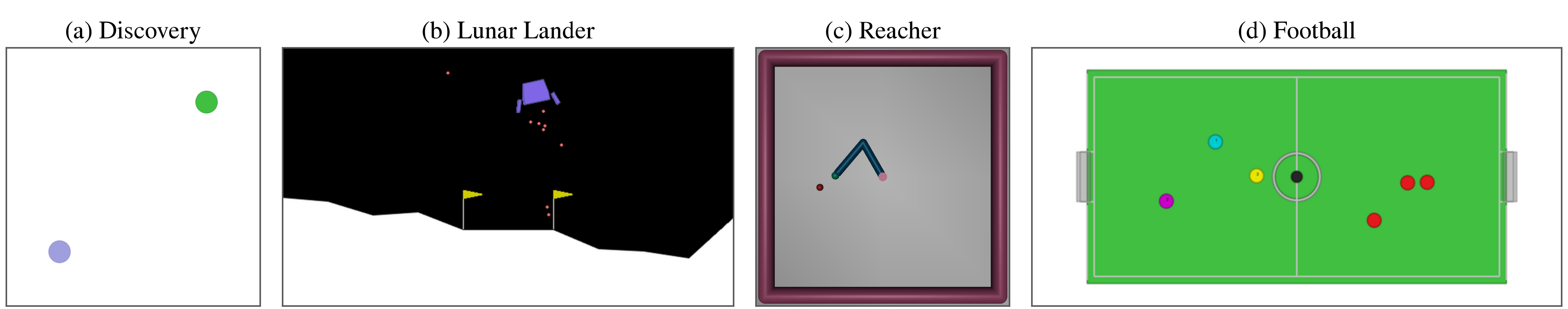}
\caption{Simulated environments. \textbf{(a)} Discovery: single-agent navigation (VMAS~\cite{bettini2022VMAS}). \textbf{(b)} Lunar Lander: powered descent (Box2D~\cite{towers2024gymnasium}). \textbf{(c)} Reacher: 2-DOF planar robotic arm (MuJoCo~\cite{todorov2012mujoco}). \textbf{(d)} Football: 3v3 multi-agent team play (VMAS~\cite{bettini2022VMAS}).}
\label{fig:environments}
\end{figure*}

\begin{figure*}[!t]
\centering
\includegraphics[width=0.86\textwidth]{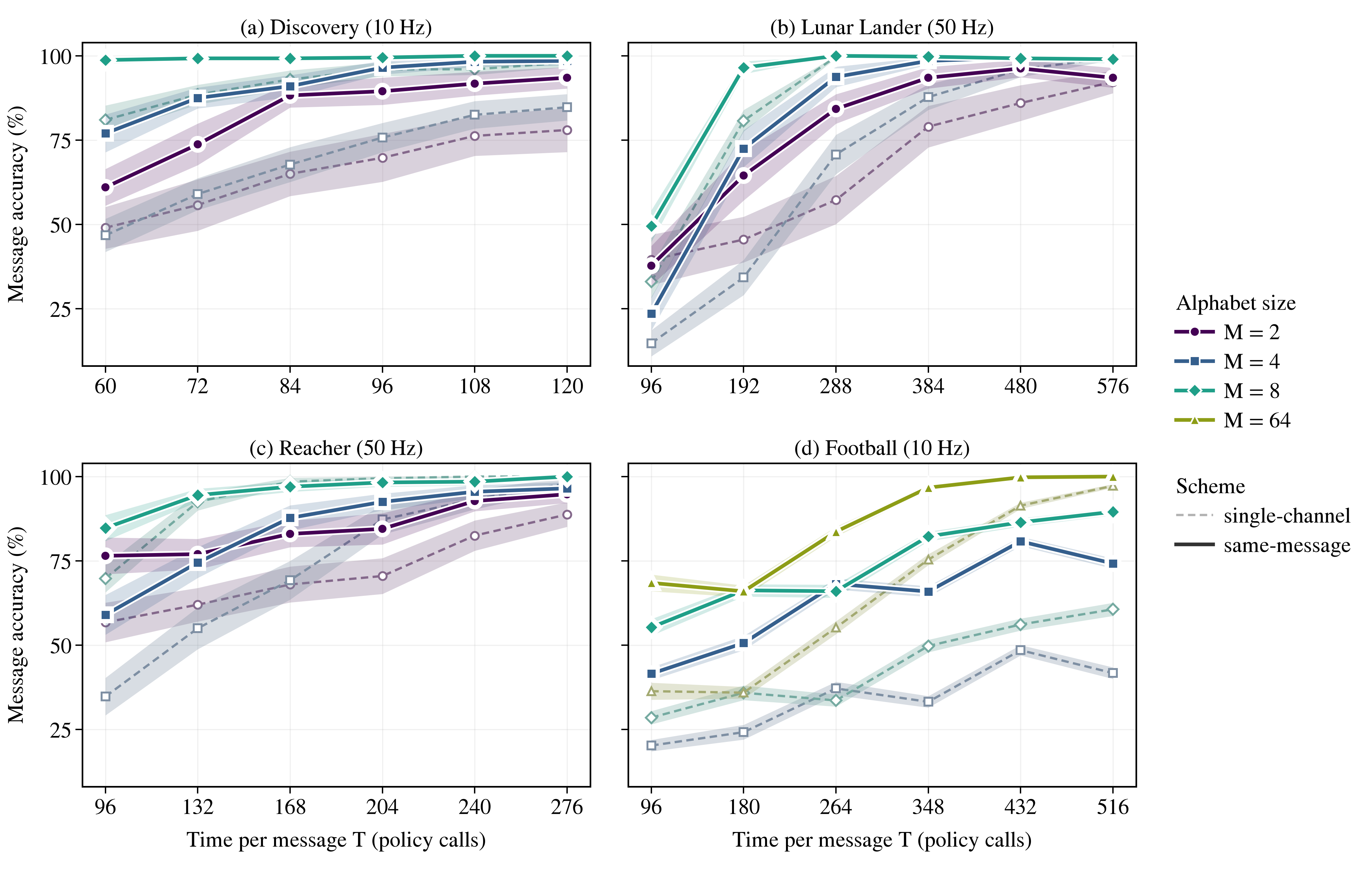}
\caption{Decoding accuracy from video across $50$ replications vs.\ alphabet size $M$ and time per message $T$: \textbf{(a)} Discovery, \textbf{(b)} Lunar Lander, \textbf{(c)} Reacher, \textbf{(d)} Football. Solid: same-message multi-channel; dashed: single-channel baseline (Discovery: $x$, Lunar Lander: side engine, Reacher: inner arm, Football: agent~0). Bands are $95\%$ CIs. Policy rate $10$ Hz (Discovery, Football) and $50$ Hz (Lunar Lander, Reacher).}
\label{fig:sweeps}
\end{figure*}

We evaluate MTM across four simulated environments of increasing difficulty, built in the VMAS, Box2D and MuJoCo simulators~\cite{bettini2022VMAS, towers2024gymnasium, todorov2012mujoco} (Fig.~\ref{fig:environments}). Decoding relies on video alone. We further validate the method on the real RoboMaster~\cite{blumenkamp2024robomaster} platform (Fig.~\ref{fig:robomaster-arena}). 
We systematically explore the design space, studying how alphabet size, channel allocation strategy, and time budget affect decoding accuracy.
Fig.~\ref{fig:sweeps} reports the single-channel and same-message schemes across alphabet size $M$ and time budget $T$; Table~\ref{tab:split} reports the split-message scheme at $M{=}4$; Fig.~\ref{fig:robomaster-plots} reports real robot results. Implementation details and full numerical results are given in the Appendix.

\subsection{Evaluation Details}

The encoded message is the agent's goal ID, or the team strategy in Football; since encoding and decoding depend only on the symbol sequence, these are illustrative payloads, not a requirement of the method. Split-message results use payloads long enough to divide across the available channels.

Each rollout uses a fresh codebook, so reported accuracies average over the codebook ensemble; in deployment, the codebook would instead be fixed, derived from a long-lived key shared by operator and auditor (App.~\ref{app:keys}). Message length, alphabet size and symbol boundaries are agreed in advance, not inferred from the recording.

Discovery, Lunar Lander, Football and RoboMaster run deterministic controllers with added Gaussian noise; Reacher runs a stochastic trained policy; MTM shapes both into the messages. Glimpses are velocities estimated by color-based tracking of video, or from motion capture on hardware.

\subsection{Main Results}

\noindent\textbf{Discovery: General Design Choices.}
The single-agent Discovery task offers a controlled, low-noise setting ideal for isolating core design choices. 
The agent navigates at $10$ Hz to one of $8$ goal locations, with the goal ID as the message (3-bit message), and the action direction ($x$ or $y$) as the channels. 
Fig.~\ref{fig:sweeps}(a) shows that accuracy increases with both alphabet size $M$ and time budget $T$. Combining both $x$ and $y$ channels under the same-message scheme consistently outperforms single-channel; notably, $M{=}8$ reaches $95\%$ accuracy across all $T$ and $100\%$ at $T{=}10.8$~s ($108$ policy calls). The two channels achieve similar individual accuracies, motivating an equal symbol allocation in the split-message scheme (Table~\ref{tab:split}), which outperforms same-message at smaller $T$.

\begin{table}[!t]
\centering
\footnotesize
\caption{Split-message: the payload is split into base-$M$ symbols ($M{=}4$
throughout) transmitted concurrently, with $(l_c)$ symbols allocated to channel
$c$. $T$ is the smallest total message time on each sweep reaching a $90\%$
message-accuracy target, in policy calls. ``Message'' is the fraction of rollouts decoding every symbol correctly.}
\label{tab:split}
\begin{tabular}{@{}llccrr@{}}
\toprule
Env & Payload & Chan. & $(l_c)$ & $T$ & Message (\%) \\
\midrule
Discovery    & 3 bit  & 2 & $(1,1)$     &  84 & 92.0 \\
Lunar Lander & 10 bit & 2 & $(2,3)$     & 576 & 90.0 \\
Reacher      & 6 bit  & 2 & $(2,1)$     & 306 & 92.5 \\
Football     & 6 bit  & 3 & $(1,1,1)$   & 432 & 90.7 \\
RoboMaster   & 8 bit  & 4 & $(1,1,1,1)$ & 300 & 95.0 \\
\bottomrule
\end{tabular}
\end{table}

\noindent\textbf{Lunar Lander: Challenging Dynamics.}
We use the Lunar Lander environment to investigate watermark robustness under challenging dynamics. The agent must land a rocket at one of $8$ possible landing pads;  its control loop actively combats noisy fluctuations, interfering with the watermark. 
The agent operates at $50$ Hz, with the landing pad ID serving as the message (3-bit message), and the action directions (side engine and main engine) as channels. 
Fig.~\ref{fig:sweeps}(b) shows that reliable decoding remains achievable despite the dynamics, with the $M{=}8$ same-message scheme reaching $100\%$ accuracy at $T{=}5.76$~s ($288$ policy calls). The $y$-channel (main engine) achieves higher accuracy at lower $T$: the environment's weaker threshold on the main engine causes less watermark clipping, making it a stronger channel. This asymmetry motivates an unequal symbol allocation in the split-message scheme (Table~\ref{tab:split}), favoring the stronger $y$-channel, which reaches $90\%$ accuracy at smaller $T$.

\begin{figure*}[!t]
\centering
\begin{subfigure}{0.35\textwidth}\centering
  \includegraphics[width=\linewidth]{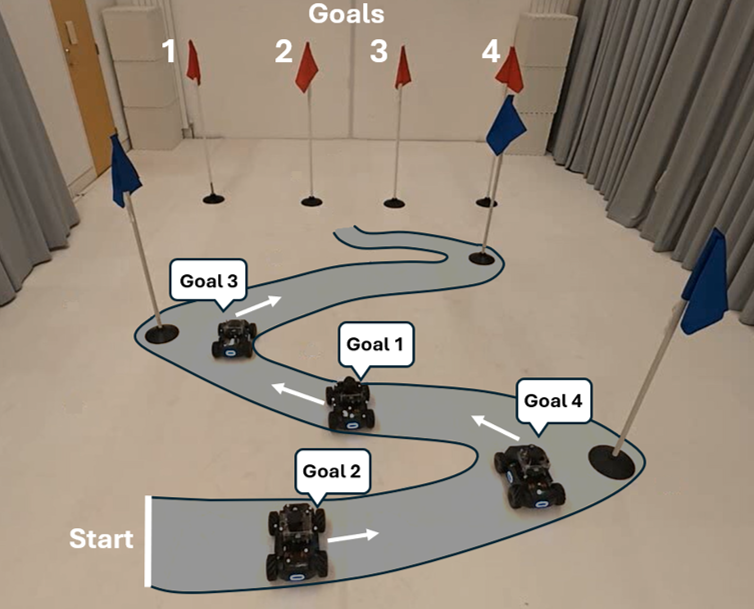}
  \caption{}\label{fig:robomaster-arena}
\end{subfigure}\hfill
\begin{subfigure}{0.6\textwidth}\centering
  \includegraphics[width=\linewidth]{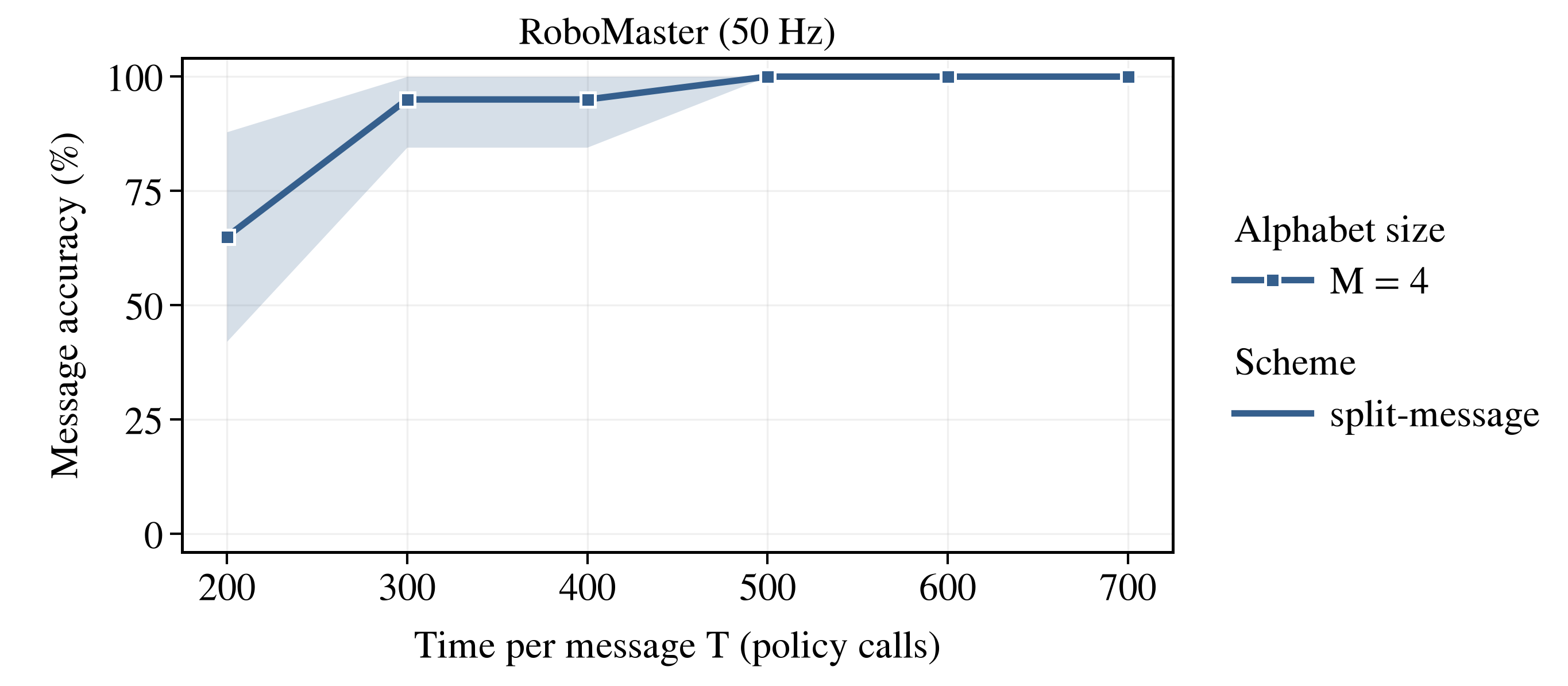}
  \caption{}\label{fig:robomaster-sweep}
\end{subfigure}
\caption{Real-world RoboMaster deployment. \textbf{(a)} The arena, each robot is an independent emitter transmitting its own message. \textbf{(b)} Decoding accuracy from motion capture across $20$ replications vs.\ time per message $T$, split-message scheme, four robots. Bands are $95\%$ CIs. Policy rate $50$ Hz. }
\label{fig:robomaster-plots}
\end{figure*}

\noindent\textbf{Reacher: Articulated Motion.}
We use the Reacher environment to investigate watermark robustness under articulated motion: angular velocity, estimated from video, is a nonlinear function of joint torques and harder to track, and the two linked arms can interfere with each other. The agent reaches one of $8$ goal positions at $50$ Hz, with the goal ID as the message (3-bit message), and the joints (inner or outer), whose torque carries the watermark, as the channels. 
Fig.~\ref{fig:sweeps}(c) shows that pooling both arms under the same-message scheme beats the single inner channel across most of the grid, decisively at small $M$ and short $T$; only at $M{=}8$ and long $T$ does the single channel stay ahead, reaching $100\%$ accuracy at $T{=}4.8$~s ($240$ calls) against $T{=}5.52$~s ($276$ calls) for same-message. The inner arm is more accurate at long symbol windows, so the split-message scheme (Table~\ref{tab:split}) favors it: allocating it two of three symbols reaches the $90\%$ target at $T{=}6.12$~s ($306$ calls) with $M{=}4$.

\noindent\textbf{Football: Multi-Agent and Adversarial.}
We use the Football environment to simultaneously push our method into three more challenging settings. First, this task is \emph{multi-agent}: three agents, controlled at $10$ Hz, play on a team, each serving as a separate channel. 
Second, it is \emph{adversarial}: an opposing team interferes with the agents' trajectories, adding collision noise that makes the team strategy difficult to recover from video. 
Third, the message is \emph{higher-level intent}: one of $64$ team strategies (6-bit message), defined by four behavioral parameters, rather than the simple goal IDs used previously. 
Fig.~\ref{fig:sweeps}(d) shows that the same-message scheme reaches $100\%$ accuracy at $M{=}64$ and $T{=}51.6$~s ($516$ policy calls). Since the three agents contribute similarly, we allocate symbols equally in the split-message scheme (Table~\ref{tab:split}), which outperforms the same-message scheme at smaller $M$, reaching $90\%$ accuracy in $T{=}43.2$~s ($432$ calls). 

\noindent\textbf{RoboMaster: Real-World Deployment.}
We validate on physical hardware with four RoboMaster robots~\cite{blumenkamp2024robomaster}, controlled at $50$ Hz, each acting as an independent emitter transmitting which of $4$ goal locations it navigates to (2-bit message). A motion capture system serves as the auditor's remote-sensing feed. Despite real-world noise and dynamics, all four robots decode correctly in every rollout under the split-message scheme at $T{=}10$~s ($500$ policy calls; Fig.~\ref{fig:robomaster-sweep})---$26\%$ of the trajectory length, well before the robots reach their goals. The $90\%$ target is met already at $T{=}6$~s ($300$ calls) (Table~\ref{tab:split}), with $95\%$ joint recovery. Over $n{=}20$ rollouts the exact $95\%$ binomial interval on joint recovery is $[83.2,\,100]\%$ (per-symbol $[95.5,\,100]\%$). The four 2-bit symbols concatenate to 8 bits in $10$~s: an aggregate $0.8$ bits/s across the team, or $0.2$ bits/s per robot. This demonstrates the practical value of the method: an auditor recovers each robot's intent well before its behavior reveals it.

\noindent\textbf{Results Summary.}
Figs.~\ref{fig:sweeps} and~\ref{fig:robomaster-sweep} suggest three design heuristics: (i) larger $M$ often improves accuracy for a fixed payload and time budget, as a larger alphabet needs fewer symbols to encode the same payload, allowing longer averaging windows; (ii) same-message encoding is the strongest general-purpose scheme, pooling score across channels; (iii) split-message encoding reduces transmission time when channels have asymmetric accuracies, as unequal symbol allocation lets stronger channels carry more of the message.
The Lunar Lander and Reacher results show why these choices cannot be made blindly: channels can differ substantially in quality, so the measured curves should set both the operating point and the allocation.

\subsection{Null Conditions}
\label{app:null}

\begin{table*}[t]
\centering
\footnotesize
\caption{Null-condition symbol accuracy (\%) at each environment's $(M, T)$
operating point, against the $1/M$ chance rate.}
\label{tab:null}
\begin{tabular}{@{}lccrrrlrrl@{}}
\toprule
& & & & \multicolumn{3}{c}{Wrong key} & \multicolumn{3}{c}{No message} \\
\cmidrule(lr){5-7}\cmidrule(lr){8-10}
Env & $M$ & $T$ & $1/M$ (\%) & $n_{\mathrm{dec}}$ & Acc. & 95\% CI
    & $n_{\mathrm{dec}}$ & Acc. & 95\% CI \\
\midrule
Discovery    & 8  & 108 & 12.50 & 2000  & 12.10 & [10.60, 13.60] & 400 & 11.75 & [8.50, 15.25] \\
Lunar Lander & 8  & 288 & 12.50 & 1000  & 14.70 & [12.60, 16.90] & 200 & 8.50  & [5.00, 12.50] \\
Reacher      & 8  & 240 & 12.50 & 2000  & 13.60 & [12.20, 15.05] & 400 & 10.50 & [7.75, 13.50] \\
Football     & 64 & 516 & 1.56  & 16000 & 1.46  & [1.28, 1.65]   & 400 & 1.50  & [0.50, 2.75] \\
RoboMaster   & 4  & 500 & 25.00 & 1600  & 24.81 & [22.69, 26.88] & 80  & 21.25 & [13.75, 28.75] \\
\bottomrule
\end{tabular}
\end{table*}

\begin{table*}[t]
\centering
\footnotesize
\caption{Deterministic policy vs.\ stochastic vs.\ MTM, 50 rollouts per env, each
with an independent random seed. Values are mean $\pm$ standard deviation over
rollouts. Bold marks the best result in each row.
``Noise only'' adds Gaussian noise of the same amplitude as MTM's that carries
no message, isolating the cost of adding stochasticity to the policy from the
cost of carrying a message. Reacher's policy is already stochastic, so it has
no deterministic condition. }

\label{tab:reward-impact}
\begin{tabular}{@{}llccc@{}}
\toprule
Env & Metric & Deterministic & Noise only & MTM \\
\midrule
Discovery     & final dist.\ to goal (m) & $0.2404 \pm 0.0479$ & $0.2420 \pm 0.0499$ & $\mathbf{0.2403} \pm 0.0481$ \\
Lunar Lander  & episodic return          & $\mathbf{265.8} \pm 85.2$ & $254.8 \pm 100.9$ & $253.8 \pm 101.6$ \\
Reacher       & 50-step return           & ---                 & $-4.31 \pm 1.21$    & $\mathbf{-4.15} \pm 1.23$ \\
Football      & mean dist.\ to goal      & $\mathbf{0.818} \pm 0.177$ & $0.984 \pm 0.252$ & $0.917 \pm 0.227$ \\
\bottomrule
\end{tabular}
\end{table*}

Table~\ref{tab:null} reports symbol accuracy under two null conditions. In the wrong-key condition, the decoder is supplied decoy codebooks regenerated from seeds unrelated to the one used at injection, with several decoys per rollout.
In the no-message condition, the policy runs without watermark injection and the
decoder is supplied the correct codebook. $n_{\mathrm{dec}}$ is the total number of individual decoded symbols. For wrong-key, each rollout is decoded against $5$ decoy codebooks independently drawn, giving larger $n_{\mathrm{dec}}$ (for RoboMaster, $20$ decoys per replication, across the four robots).

Recovery at or near chance establishes that decoding requires the key and that
the decoder does not report spurious symbols from unwatermarked motion. 
These results characterize robustness to key-guessing and false-positive detection; concealment against steganalysis, repeated-message analysis, or classifier-based detection remains open for future work.

\subsection{Task Performance Preservation}
\label{app:reward-impact}

We compare reward distributions across three conditions: the policy as
deterministic; the stochastic policy with white Gaussian noise at MTM's amplitude,
which carries no message; and the same policy carrying a message with MTM,
whose noise is colored. The middle condition separates the cost of adding stochasticity to the policy at all from the cost of shaping that stochasticity into a message.

For each environment, we report the following task-specific metrics (exact definitions in App.~\ref{app:reward-defs}). In Discovery, we use distance to the goal. Lunar Lander
uses Box2D's standard reward, corresponding to distance to the landing pad with
an energy penalty for engine use. Reacher uses MuJoCo's reward, based on
distance of the end effector to the target with an energy penalty. In
Football, we use the mean distance from the ball to the attacking goal and the scoring rate.

Table~\ref{tab:reward-impact} shows no detectable effect in Discovery, Lunar
Lander or Reacher. In Football, using a stochastic policy reduces scoring rate ($1.00 \to 0.88$; MTM $0.90$) and
increases mean ball-to-goal distance ($0.818 \to 0.984$; MTM $0.917$); MTM adds no further cost beyond this, confirming that the cost comes from stochasticity rather than MTM itself. 
We also measured tracking error, with no measurable difference from deterministic rollouts.

\begin{table}[!t]
\centering
\footnotesize
\caption{Attacks on Discovery ($M{=}8$, $T{=}120$) over $n{=}50$ rollouts. Additive-noise strength is given as a multiple of the watermark's standard deviation. $\Delta$cmd is the root-mean-square (RMS) change to the commanded
action, relative to its unattacked RMS; $\Delta$dist is the increase
in mean agent--goal distance. Accuracy chance rate is $12.5\%$.}
\label{tab:attacks}

\begin{tabular}{@{}rrrr@{}}
\toprule
\multicolumn{4}{@{}l}{\emph{Additive noise}} \\
$\sigma_{\mathrm{adv}}/\sigma_{\mathrm{wm}}$ & Msg.\ acc.\ (\%) & $\Delta$cmd (\%) & $\Delta$dist (\%) \\
\midrule
0.00 & 100.0 &   0.0 & 0.0 \\
0.83 &  98.5 &  18.1 & 0.0 \\
1.67 &  67.0 &  36.2 & 0.1 \\
3.33 &  24.5 &  72.4 & 0.3 \\
5.00 &  16.5 & 108.5 & 0.5 \\
\bottomrule
\end{tabular}

\vspace{4pt}

\begin{tabular}{@{}rrrr@{}}
\toprule
\multicolumn{4}{@{}l}{\emph{Band-stop filtering}} \\
Order & Msg.\ acc.\ (\%) & $\Delta$cmd (\%) & $\Delta$dist (\%) \\
\midrule
0  & 100.0 &  0.0 &  0.0 \\
2  &  22.5 & 36.2 &  4.3 \\
6  &  12.3 & 53.6 & 12.2 \\
10 &  12.0 & 66.5 & 21.5 \\
\bottomrule
\end{tabular}

\end{table}

\subsection{Adversarial Robustness}
\label{app:adversarial}

We consider an adversary who has obtained the watermarked policy and wishes to
suppress the embedded message while preserving task utility. The adversary can
modify commanded actions before execution and collect offline rollouts, but
does not hold the key $k$ and cannot tamper with the auditor's sensing. For the
band-stop attack, we additionally assume the secret band $B$ is known. Full attack sweeps are in App.~\ref{app:attacks-full}.

\noindent\textbf{Additive noise.}
The adversary adds Gaussian noise
$\eta_{\mathrm{adv}} \sim \mathcal{N}(0, \sigma_{\mathrm{adv}}^2 I)$ to the
policy's actions before execution, lowering the
signal-to-interference-plus-noise ratio (SINR) in the watermark band. By the
SINR--coherency relation of CoNoCo~\cite{conoco}, raising the noise power
spectral density $P_N(f)$ lowers the squared coherency $|C_{WG}(f)|^2 =
\mathrm{SINR}/(\mathrm{SINR}{+}1)$ between watermark and glimpse, reducing the
per-symbol score margin and hence accuracy $a_c$ (Table~\ref{tab:attacks}).
The attacker pays in command effort, not task performance: driving accuracy near chance costs over $100\%$ RMS command distortion but under $1\%$ goal-distance
degradation, since zero-mean noise averages out over Discovery's slow
goal-seeking dynamics.

\noindent\textbf{Band-stop filtering.}
An adversary with knowledge of the secret band $B$ could instead filter this frequency band directly.  A band-stop notch over $B$ attenuates the watermark, but also
removes any policy content in $B$. With $B$ chosen to overlap the policy's
dominant frequencies, an order-2 Butterworth filter drops message accuracy from
$100\%$ to $22.5\%$ but inflicts $36\%$ command-space distortion (RMS post-
vs.\ pre-attack action) and a $4.3\%$ increase in mean agent--goal distance;
driving accuracy to chance requires order 6, at $54\%$ distortion and
$12.2\%$. 
Unlike additive noise, band-stop filtering suppresses the watermark at moderate cost, making it the more practical attack against our approach.

\noindent\textbf{Other Potential Attacks.}
Beyond additive noise and band-stop filtering, richer adversarial strategies are an interesting direction for future work. For instance, because the watermark has structure across time rather than being purely random, a learned jammer could in principle exploit that structure from observed motion alone, without recovering the key. Another direction is \emph{desynchronization}, deliberately misaligning the timing of actions to disrupt symbol boundaries during decoding.


\section{Conclusion}
\label{sec:conclusion}

We introduced \emph{Messaging Through Motion} (MTM), a method for transmitting keyed messages through a robot's physical movement. MTM makes motion itself a keyed communication channel: each message symbol is mapped to a noise signal, injected through a pre-trained stochastic policy's action-noise channel at inference time, and decoded from remote motion observations by an auditor holding the shared key. The method requires no retraining, no architectural modification, and no additional hardware, so the channel is open to any robot that moves and can be seen.

Across four simulated domains and a real-robot RoboMaster deployment, MTM shows that motion can carry useful message content while preserving the robot’s primary behavior. 
The experiments expose the main design trade-offs: (i) larger alphabets reduce total transmission time, (ii) same-message encoding improves robustness, and (iii) channels can differ substantially in reliability.

More broadly, this work demonstrates that subtle statistical signals in robot motion can serve as a communication medium. Unlike conventional robot communication, which depends on explicit communication links or onboard logs access, motion is already observable whenever a robot acts. A keyed motion channel can therefore complement existing communication systems, giving auditors, safety monitors and regulators a way to recover intent, identity, or protocol information from physical behavior alone.

\section{Limitations}
\label{sec:limitations}

MTM applies where messages are discrete and piecewise constant over a symbol
window. Per-symbol errors compound multiplicatively across a message, so the
payloads we demonstrate are short; extending reliably to longer messages, with
error correction across symbols, is the natural next step.

MTM hides message content but not the presence of a watermark: as noted for
CoNoCo~\cite{conoco}, colored noise preserves each action's marginal distribution but not its temporal
correlation, so an observer without the key could still detect that a watermark
is present. This would require recordings together with a reference
for the robot's unwatermarked motion, to reveal excess power in a particular frequency band,
or, with a fixed codebook, multiple recordings of the same symbol being sent, to
reveal correlation between them.
Additionally, MTM proves that a message came from someone holding the key, but
not who that sender was; adding an authentication layer above MTM is a
promising direction.

MTM also does not prevent replay attacks: nothing binds a transmission to a
time, so a recording of a valid message (e.g.\ resubmitted footage) decodes as
valid again. A time-varying codebook, derived from the key and a timestamp,
would prevent this.

Our experimental results also assume the auditor knows when the message begins.
This assumption could be relaxed using GCC-PHAT, a standard technique for
estimating time offsets between two signals via cross-correlation, as
demonstrated in CoNoCo~\cite{conoco} for recovering the offset between a
policy's execution clock and an observer's sampling; we leave this to future
work.

\section*{Acknowledgment}
This work is supported by European Research Council (ERC) Project 949940 (gAIa), and by the EPSRC funded INFORMED-AI project EP/Y028732/1. We gratefully acknowledge their support.
Claude was used, under the authors' review, to write and debug parts of the experiment code and the plotting scripts that produce the results in Sec.~\ref{sec:results} and the Appendix.

\bibliographystyle{IEEEtran}
\bibliography{IEEEabrv,references}

\onecolumn
\section*{Appendix}
\setcounter{subsection}{0}
\renewcommand{\thesubsection}{\Alph{subsection}}
\makeatletter
\renewcommand{\p@subsection}{}
\makeatother

\subsection{Notation}
\label{app:notation}

\begin{table}[H]
\centering
\caption{Key mathematical symbols used in this paper.}
\label{tab:symbols}
\begin{tabular}{l p{0.72\linewidth}}
\toprule
Symbol & Description \\
\midrule
\multicolumn{2}{l}{\textit{Problem setting}} \\
$\pi$ & Pre-trained stochastic policy \\
$m$ & Message transmitted to the auditor \\
$L$ & Message length (number of symbols) \\
$M$ & Alphabet size (symbols per window) \\
$k$ & Shared secret key \\
$T$ & Total transmission time \\
$\epsilon$ & Decoding error tolerance \\
\midrule
\multicolumn{2}{l}{\textit{Glimpse / physical observation}} \\
$\{T_k\}$ & Policy execution times \\
$a_{\mathrm{exec}}(t)$ & Executed action \\
$S_{\mathrm{dyn}}$ & State dynamics \\
$\{t_i\},\,f_g$ & Remote sensor sampling times and rate \\
$\mathcal{G} = (G_i)_{i=0}^{N-1}$ & Glimpse sequence (auditor's sole data) \\
$G_{\mathrm{map}}$ & State $\to$ remote observation map \\
$\eta_i$ & Measurement noise on glimpse $G_i$ \\
$B$ & Secret frequency band for CGN injection \\
\midrule
\multicolumn{2}{l}{\textit{Encoding / decoding}} \\
$k_i$ & Per-symbol keyed seed for symbol $i$ \\
$c$ & Channel index (observable degree of freedom) \\
$\mathcal{I}$ & Subset of channels used for transmission \\
$l_c$ & Number of symbols allocated to channel $c$ \\
$a_c(t, M)$ & Empirical per-symbol accuracy on channel $c$ \\
$a_{\mathcal{I}}(t, M)$ & Pooled per-symbol accuracy over $\mathcal{I}$ \\
\bottomrule
\end{tabular}
\end{table}

\subsection{Policy and Controller Details}
\label{app:policies}

For all environments, each rollout uses a distinct codebook
$\bigl(\text{wm\_master\_seed} = \text{base} + i\bigr)$ paired with a
distinct env seed, so the 50 rollouts per $(M, T)$ cell are i.i.d.\
draws from the joint (codebook, env-seed) distribution. This removes
single-codebook bias from the reported accuracies. It is a
variance-reduction measure rather than part of the protocol; see
App.~\ref{app:keys}.

\noindent\textbf{Discovery.}
A scripted navigation controller steers the agent toward the current
goal at unit speed at $10$\,Hz. Watermarks are injected as additive CGN
perturbations on the 2D velocity command ($\sigma{=}0.03$). The agent
visits $8$ sequential goals per rollout; each goal encodes a 3-bit goal
index. Results are averaged over $50$ independently seeded rollouts
per $(M, T)$ setting, each with its own codebook. Video is recorded at
$704{\times}704$\,px.

\noindent\textbf{Lunar Lander.}
A PD hover controller guides the lander toward the target pad at
$50$\,Hz. A 150-step off-camera warmup lets the lander descend into
the trackable viewport before watermark injection and video recording
begin. $\sigma{=}0.5$; 8 goals per rollout, $50$ independently seeded
rollouts per $(M, T)$ setting, each with its own codebook. Video is
recorded at $600{\times}400$\,px.

\noindent\textbf{Reacher.}
SAC (Stable-Baselines3, seed 42) trained for $1.9 \times 10^5$ steps
on a modified Reacher with target positions sampled within
$\pm 0.20$ in each axis, controlled at $50$\,Hz. During injection the
arm visits 8 sequential targets on a ring of radius $r{=}0.14$, each
encoding a 3-bit sector index. $\sigma{=}1.0$; $50$ independently
seeded rollouts per $(M, T)$ setting, each with its own codebook.
Video is recorded at $480{\times}480$\,px.

\noindent\textbf{Football.}
A heuristic PD controller drives a team of three agents in a 3v3
adversarial setting at $10$\,Hz ($\Delta t{=}0.1$\,s). 64 team
strategies are generated by varying four behavioral parameters
(tempo, approach bias, formation, passing frequency). All three
agents carry the same CGN pattern on their 2D action vectors
($\sigma{=}0.3$). For each $(M, T)$ cell every one of the 64
strategies is evaluated, each with $50$ rollouts using independent
$(\text{codebook},\,\text{env-seed})$ pairs (3{,}200 rollouts per
cell). Video is recorded at $1200{\times}800$\,px (VMAS default),
rendered to an invisible pyglet window to avoid display-context
contention between parallel workers.

\subsection{Detection Pipeline Details}
\label{app:detection}

All environments share a common coherency-based detector. Each rollout
regenerates the CGN reference table from the rollout's
\texttt{wm\_master\_seed} (the same seed used at injection time), so
detection sees a different codebook on every rollout.
Cross-spectral densities are estimated via Welch's method with a Hann
window, segment length
$\mathrm{nperseg}{=}\min\!\bigl(128,\,\lfloor L/2 \rfloor\bigr)$,
overlap $\tfrac{3}{4}\times\mathrm{nperseg}$, and
$\mathrm{nfft}{=}L$ to align Welch frequency bins with CGN
injection frequencies. The watermark band is
$[0.5,\,5.0]$\,Hz for Lunar Lander and Reacher,
$[0.10,\,0.20]$\,Hz for Discovery, and
$[0.10,\,0.35]$\,Hz for Football
(the latter two normalized, $f_s{=}1.0$).
A grid of 101 candidate frequency scales over
$[0.8\,f_{\pi},\,1.2\,f_{\pi}]$ is searched to account for
synchronization uncertainty between the policy execution rate and
the observer's sampling rate.

\noindent\textbf{Tracking methods.}
Discovery uses color-centroid tracking.
Lunar Lander uses brightness-centroid tracking of bright pixels in a
region of interest.
Reacher uses HSV color segmentation to track the inner- and outer-arm centroids
from the top-down camera view; the inner-joint rate is the derivative of the
inner-link angle, and the outer-joint rate the derivative of the outer-link angle
minus the inner-link angle.
Football uses HSV color segmentation (cyan/magenta/yellow agents
at the native $1200{\times}800$ render; velocity via finite differences
with spike removal to suppress goal-reset teleport artifacts).
Tracked positions in Lunar Lander are smoothed with a
constant-velocity Kalman filter (process noise std $= 6.0$,
measurement noise std $= 0.8$).

\subsection{Codebook Construction}
\label{app:codebook}

Attribution errors arise when multiple hypotheses yield similar
coherency scores on the same glimpse sequence.

One mitigation is using different frequency bands per symbol, though
this is constrained by available bandwidth and the maximum frequency
at which video-based detection remains reliable.

Iterative codebook construction (e.g.\ via
A-BSTA~\cite{jiang2026watermark}) could also reduce inter-symbol
confusion, but is only approximate in our setting since detection
operates between observed glimpses and CGN references rather than
between CGN sequences directly.

Orthogonal frequency-domain signals (e.g.\ Hadamard matrices) face
the same limitation: they optimize inter-signal orthogonality rather
than the magnitude coherence between observed residuals and keyed
references that our detector actually scores.

In practice, independently sampled random seeds provide sufficient
separation without explicit optimization; we additionally vary the
codebook across rollouts ($\text{wm\_master\_seed} = \text{base} + i$
for rollout $i$) so reported accuracies are averaged over the
codebook ensemble rather than fixed to a single draw. We leave
principled codebook design to future work.

\subsection{Keys and Synchronization}
\label{app:keys}

In deployment the codebook is fixed, derived from a long-lived key shared
between the robot's operator and the auditor. The per-rollout fresh codebook
used throughout our experiments (App.~\ref{app:policies}) is a
variance-reduction measure that averages reported accuracies over the codebook
ensemble; it is not part of the protocol, and our results should not be read as
evaluating a one-time-key regime.

The symbol rate and any onset offset between the policy's execution clock and
the observer's sampling are recoverable by the primitive's frequency-scale
search and GCC-PHAT alignment, as in CoNoCo~\cite{conoco}. Message length,
alphabet size and symbol boundaries are agreed in advance alongside the key, so
the receiver does not infer them from the recording. Replay and message
substitution by another key-holder are not prevented: the scheme proves
possession of the key, not message origin or integrity. Binding a message to a
sender or a time would require an authentication layer above MTM, which we
leave to future work.

\subsection{Discovery: Numerical Results}
\label{app:discovery-results}

Throughout, $\sigma$ denotes the watermark noise amplitude and $n$ the number of independently seeded rollouts per setting. Errors are reported as standard error of the mean. The Discovery environment is symmetric in $x$/$y$, so we only sweep one single-channel axis.

\begin{table}[H]
\centering
\caption{Discovery single-channel (x-dim): message accuracy (\%) vs.\ $T$
and $M$ ($\sigma{=}0.03$, 8 goals/rollout, $n{=}50$).}
\label{tab:discovery-single-x}
\begin{tabular}{r rrr rrr}
\toprule
& \multicolumn{3}{c}{Message Accuracy (\%)}
& \multicolumn{3}{c}{Margin} \\
\cmidrule(lr){2-4}\cmidrule(lr){5-7}
$T$ & $M{=}2$ & $M{=}4$ & $M{=}8$
& $M{=}2$ & $M{=}4$ & $M{=}8$ \\
\midrule
48  & $43.3{\scriptstyle\pm3.4}$ & $33.8{\scriptstyle\pm2.6}$ & $79.0{\scriptstyle\pm1.8}$ & 0.09 & 0.05 & 0.04 \\
60  & $49.0{\scriptstyle\pm3.1}$ & $46.8{\scriptstyle\pm2.4}$ & $81.0{\scriptstyle\pm2.1}$ & 0.11 & 0.06 & 0.05 \\
72  & $55.8{\scriptstyle\pm3.8}$ & $59.0{\scriptstyle\pm2.3}$ & $88.5{\scriptstyle\pm1.5}$ & 0.11 & 0.05 & 0.06 \\
84  & $65.0{\scriptstyle\pm3.3}$ & $67.8{\scriptstyle\pm2.6}$ & $93.0{\scriptstyle\pm1.3}$ & 0.11 & 0.07 & 0.06 \\
96  & $69.8{\scriptstyle\pm3.5}$ & $75.8{\scriptstyle\pm2.2}$ & $96.0{\scriptstyle\pm1.0}$ & 0.12 & 0.06 & 0.07 \\
108 & $76.3{\scriptstyle\pm2.9}$ & $82.5{\scriptstyle\pm2.0}$ & $96.0{\scriptstyle\pm0.8}$ & 0.12 & 0.09 & 0.07 \\
120 & $78.0{\scriptstyle\pm3.3}$ & $84.8{\scriptstyle\pm1.9}$ & $98.0{\scriptstyle\pm0.6}$ & 0.13 & 0.07 & 0.09 \\
\bottomrule
\end{tabular}
\end{table}

\begin{table}[H]
\centering
\caption{Discovery multi-channel same-message: message accuracy (\%)
($\sigma{=}0.03$, 8 goals/rollout, $n{=}50$).}
\label{tab:discovery-same}
\begin{tabular}{r rrr rrr}
\toprule
& \multicolumn{3}{c}{Message Accuracy (\%)}
& \multicolumn{3}{c}{Margin} \\
\cmidrule(lr){2-4}\cmidrule(lr){5-7}
$T$ & $M{=}2$ & $M{=}4$ & $M{=}8$
& $M{=}2$ & $M{=}4$ & $M{=}8$ \\
\midrule
48  & $49.0{\scriptstyle\pm2.7}$ & $55.8{\scriptstyle\pm3.3}$ & $96.5{\scriptstyle\pm0.9}$ & 0.10 & 0.07 & 0.07 \\
60  & $61.0{\scriptstyle\pm2.7}$ & $77.0{\scriptstyle\pm2.8}$ & $98.8{\scriptstyle\pm0.5}$ & 0.13 & 0.09 & 0.09 \\
72  & $73.8{\scriptstyle\pm3.1}$ & $87.5{\scriptstyle\pm1.5}$ & $99.3{\scriptstyle\pm0.4}$ & 0.13 & 0.09 & 0.10 \\
84  & $88.3{\scriptstyle\pm1.8}$ & $91.0{\scriptstyle\pm1.6}$ & $99.3{\scriptstyle\pm0.4}$ & 0.15 & 0.12 & 0.11 \\
96  & $89.5{\scriptstyle\pm2.0}$ & $96.5{\scriptstyle\pm1.0}$ & $99.5{\scriptstyle\pm0.3}$ & 0.15 & 0.11 & 0.12 \\
108 & $91.8{\scriptstyle\pm1.8}$ & $98.3{\scriptstyle\pm0.8}$ & 100.0                       & 0.16 & 0.13 & 0.12 \\
120 & $93.5{\scriptstyle\pm1.6}$ & $98.5{\scriptstyle\pm0.7}$ & 100.0                       & 0.16 & 0.12 & 0.13 \\
\bottomrule
\end{tabular}
\end{table}

\begin{table}[H]
\centering
\caption{Discovery multi-channel split-message: message accuracy
(\%) at $M{=}4$, $L{=}2$ symbols per goal, allocation $(l_x, l_y){=}(1,1)$
($\sigma{=}0.03$, 8 goals/rollout, $n{=}50$).}
\label{tab:discovery-indep}
\begin{tabular}{r r r}
\toprule
$T$ & Msg.\ Acc.\ (\%) & Margin \\
\midrule
48  & $75.8{\scriptstyle\pm2.4}$ & 0.06 \\
60  & $83.5{\scriptstyle\pm2.0}$ & 0.08 \\
72  & $89.5{\scriptstyle\pm1.4}$ & 0.08 \\
84  & $92.0{\scriptstyle\pm1.4}$ & 0.09 \\
96  & $95.8{\scriptstyle\pm1.3}$ & 0.10 \\
108 & $95.8{\scriptstyle\pm1.2}$ & 0.10 \\
120 & $98.0{\scriptstyle\pm0.7}$ & 0.11 \\
\bottomrule
\end{tabular}
\end{table}

\subsection{Lunar Lander: Numerical Results}
\label{app:lunar-lander-results}

\begin{table}[H]
\centering
\caption{Lunar Lander single-channel (x-dim / lateral thrust): message accuracy (\%) vs.\ $T$
  and $M$ ($\sigma{=}0.5$, 8 goals/rollout, $n{=}50$).}
\label{tab:lunarlander-single-x}
\begin{tabular}{r rrr rrr}
\toprule
& \multicolumn{3}{c}{Message Accuracy (\%)}
& \multicolumn{3}{c}{Margin} \\
\cmidrule(lr){2-4}\cmidrule(lr){5-7}
$T$ & $M{=}2$ & $M{=}4$ & $M{=}8$
    & $M{=}2$ & $M{=}4$ & $M{=}8$ \\
\midrule
96  & $39.5{\scriptstyle\pm3.7}$ & $14.8{\scriptstyle\pm1.9}$ & $33.0{\scriptstyle\pm2.4}$ & 0.12 & 0.07 & 0.04 \\
192 & $45.5{\scriptstyle\pm3.4}$ & $34.3{\scriptstyle\pm2.6}$ & $80.8{\scriptstyle\pm1.6}$ & 0.12 & 0.06 & 0.06 \\
288 & $57.3{\scriptstyle\pm3.6}$ & $70.8{\scriptstyle\pm2.9}$ & $99.8{\scriptstyle\pm0.2}$ & 0.11 & 0.08 & 0.13 \\
384 & $79.0{\scriptstyle\pm3.1}$ & $87.8{\scriptstyle\pm1.9}$ & 100.0                       & 0.12 & 0.10 & 0.23 \\
480 & $86.0{\scriptstyle\pm2.6}$ & $96.0{\scriptstyle\pm1.1}$ & 100.0                       & 0.13 & 0.12 & 0.28 \\
576 & $92.3{\scriptstyle\pm1.7}$ & $99.5{\scriptstyle\pm0.3}$ & $99.8{\scriptstyle\pm0.2}$ & 0.14 & 0.15 & 0.31 \\
\bottomrule
\end{tabular}
\end{table}

\begin{table}[H]
\centering
\caption{Lunar Lander single-channel (y-dim / main engine): message accuracy (\%) vs.\ $T$
  and $M$ ($\sigma{=}0.5$, 8 goals/rollout, $n{=}50$).}
\label{tab:lunarlander-single-y}
\begin{tabular}{r rrr rrr}
\toprule
& \multicolumn{3}{c}{Message Accuracy (\%)}
& \multicolumn{3}{c}{Margin} \\
\cmidrule(lr){2-4}\cmidrule(lr){5-7}
$T$ & $M{=}2$ & $M{=}4$ & $M{=}8$
    & $M{=}2$ & $M{=}4$ & $M{=}8$ \\
\midrule
96  & $23.3{\scriptstyle\pm2.1}$ & $12.8{\scriptstyle\pm2.0}$ & $31.5{\scriptstyle\pm2.3}$ & 0.13 & 0.07 & 0.04 \\
192 & $43.3{\scriptstyle\pm3.3}$ & $56.3{\scriptstyle\pm3.4}$ & $93.3{\scriptstyle\pm1.1}$ & 0.13 & 0.08 & 0.10 \\
288 & $78.3{\scriptstyle\pm2.8}$ & $89.5{\scriptstyle\pm1.6}$ & $99.3{\scriptstyle\pm0.4}$ & 0.13 & 0.11 & 0.17 \\
384 & $89.0{\scriptstyle\pm1.9}$ & $94.3{\scriptstyle\pm1.3}$ & $99.5{\scriptstyle\pm0.3}$ & 0.17 & 0.13 & 0.27 \\
480 & $94.0{\scriptstyle\pm1.5}$ & $96.0{\scriptstyle\pm1.1}$ & $99.0{\scriptstyle\pm0.5}$ & 0.17 & 0.15 & 0.33 \\
576 & $90.8{\scriptstyle\pm2.0}$ & $96.3{\scriptstyle\pm1.0}$ & $98.0{\scriptstyle\pm0.6}$ & 0.18 & 0.19 & 0.36 \\
\bottomrule
\end{tabular}
\end{table}

\begin{table}[H]
\centering
\caption{Lunar Lander multi-channel same-message: message accuracy
  (\%) ($\sigma{=}0.5$, 8 goals/rollout, $n{=}50$).}
\label{tab:lunarlander-same}
\begin{tabular}{r rrr rrr}
\toprule
& \multicolumn{3}{c}{Message Accuracy (\%)}
& \multicolumn{3}{c}{Margin} \\
\cmidrule(lr){2-4}\cmidrule(lr){5-7}
$T$ & $M{=}2$ & $M{=}4$ & $M{=}8$
    & $M{=}2$ & $M{=}4$ & $M{=}8$ \\
\midrule
96  & $37.8{\scriptstyle\pm2.9}$ & $23.5{\scriptstyle\pm2.5}$ & $49.5{\scriptstyle\pm2.3}$ & 0.11 & 0.06 & 0.03 \\
192 & $64.5{\scriptstyle\pm3.7}$ & $72.5{\scriptstyle\pm2.7}$ & $96.5{\scriptstyle\pm0.9}$ & 0.11 & 0.07 & 0.10 \\
288 & $84.3{\scriptstyle\pm2.2}$ & $93.8{\scriptstyle\pm1.5}$ & 100.0                       & 0.11 & 0.12 & 0.18 \\
384 & $93.5{\scriptstyle\pm1.6}$ & $98.5{\scriptstyle\pm0.6}$ & $99.8{\scriptstyle\pm0.2}$ & 0.16 & 0.13 & 0.28 \\
480 & $96.3{\scriptstyle\pm1.3}$ & $99.5{\scriptstyle\pm0.3}$ & $99.3{\scriptstyle\pm0.4}$ & 0.16 & 0.15 & 0.33 \\
576 & $93.5{\scriptstyle\pm1.5}$ & $99.5{\scriptstyle\pm0.3}$ & $99.0{\scriptstyle\pm0.6}$ & 0.16 & 0.20 & 0.36 \\
\bottomrule
\end{tabular}
\end{table}

At $M{=}2$, accuracy dips slightly at $T{=}576$ versus
$T{=}480$ for both the $y$-channel and same-message scheme.
The longer watermark requires an extended hover phase, where the control loop dominates near the goal and interferes most with the final
symbol (3 of 3 at $M{=}2$); per-symbol diagnostics confirm it carries the
bulk of the added errors.

\begin{table}[H]
\centering
\caption{Lunar Lander multi-channel split-message: empirical
  validation of the predicted optimal allocation for a 10-bit message
  ($L{=}5$ symbols) at $M{=}4$, $\sigma{=}0.5$, $n{=}50$. The
  $(l_x, l_y){=}(2,3)$ allocation was selected by the analytic
  simulator from the single-channel curves; measured message accuracy
  matches the prediction (0.90 vs 0.91 predicted).}
\label{tab:lunarlander-indep}
\begin{tabular}{r r r rr r rr r}
\toprule
$M$ & $L$ & $(l_x, l_y)$ & $t_x$ & $t_y$ & $T$
    & Msg.\ Acc.\ (\%) & Sym.\ Acc.\ (\%) & Margin \\
\midrule
4 & 5 & (2, 3) & 288 & 192 & 576
  & $90.0{\scriptstyle\pm4.2}$ & $98.0{\scriptstyle\pm0.8}$ & 0.14 \\
\bottomrule
\end{tabular}
\end{table}

\subsection{Reacher: Numerical Results}
\label{app:reacher-results}

\begin{table}[H]
\centering
\caption{Reacher single-channel (inner arm / shoulder joint): message accuracy (\%) vs.\ $T$ and
  $M$ ($\sigma{=}1.0$, 8 targets/rollout, $n{=}50$).}
\label{tab:reacher-single-inner}
\begin{tabular}{r rrr rrr}
\toprule
& \multicolumn{3}{c}{Message Accuracy (\%)}
& \multicolumn{3}{c}{Margin} \\
\cmidrule(lr){2-4}\cmidrule(lr){5-7}
$T$ & $M{=}2$ & $M{=}4$ & $M{=}8$
    & $M{=}2$ & $M{=}4$ & $M{=}8$ \\
\midrule
96  & $56.75{\scriptstyle\pm2.92}$ & $34.75{\scriptstyle\pm2.77}$ & $69.75{\scriptstyle\pm2.03}$ & 0.152 & 0.088 & 0.065 \\
128 & $63.75{\scriptstyle\pm2.46}$ & $55.75{\scriptstyle\pm3.28}$ & $91.00{\scriptstyle\pm1.46}$ & 0.210 & 0.101 & 0.111 \\
132 & $62.00{\scriptstyle\pm2.47}$ & $55.00{\scriptstyle\pm3.08}$ & $92.50{\scriptstyle\pm1.27}$ & 0.190 & 0.110 & 0.114 \\
160 & $66.00{\scriptstyle\pm2.74}$ & $66.50{\scriptstyle\pm3.11}$ & $97.50{\scriptstyle\pm0.93}$ & 0.196 & 0.121 & 0.142 \\
168 & $68.00{\scriptstyle\pm2.65}$ & $69.25{\scriptstyle\pm2.86}$ & $98.50{\scriptstyle\pm0.57}$ & 0.182 & 0.126 & 0.149 \\
192 & $74.25{\scriptstyle\pm2.61}$ & $79.75{\scriptstyle\pm2.71}$ & $99.25{\scriptstyle\pm0.42}$ & 0.189 & 0.137 & 0.165 \\
204 & $70.50{\scriptstyle\pm2.64}$ & $87.25{\scriptstyle\pm2.19}$ & $99.50{\scriptstyle\pm0.35}$ & 0.193 & 0.156 & 0.180 \\
240 & $82.50{\scriptstyle\pm2.24}$ & $93.25{\scriptstyle\pm1.47}$ & 100.00                       & 0.200 & 0.178 & 0.197 \\
276 & $88.75{\scriptstyle\pm1.81}$ & $98.00{\scriptstyle\pm0.82}$ & 100.00                       & 0.214 & 0.197 & 0.245 \\
288 & $88.50{\scriptstyle\pm1.80}$ & $97.75{\scriptstyle\pm0.84}$ & 100.00                       & 0.212 & 0.190 & 0.251 \\
\bottomrule
\end{tabular}
\end{table}

\begin{table}[H]
\centering
\caption{Reacher single-channel (outer arm / elbow joint): message accuracy (\%) vs.\ $T$
  and $M$ ($\sigma{=}1.0$, 8 targets/rollout, $n{=}50$).}
\label{tab:reacher-single-outer}
\begin{tabular}{r rrr rrr}
\toprule
& \multicolumn{3}{c}{Message Accuracy (\%)}
& \multicolumn{3}{c}{Margin} \\
\cmidrule(lr){2-4}\cmidrule(lr){5-7}
$T$ & $M{=}2$ & $M{=}4$ & $M{=}8$
    & $M{=}2$ & $M{=}4$ & $M{=}8$ \\
\midrule
96  & $61.25{\scriptstyle\pm3.09}$ & $56.00{\scriptstyle\pm2.74}$ & $80.75{\scriptstyle\pm1.94}$ & 0.143 & 0.083 & 0.091 \\
128 & $75.25{\scriptstyle\pm2.59}$ & $74.25{\scriptstyle\pm2.22}$ & $91.50{\scriptstyle\pm1.34}$ & 0.209 & 0.105 & 0.135 \\
132 & $73.50{\scriptstyle\pm2.49}$ & $72.00{\scriptstyle\pm2.25}$ & $91.50{\scriptstyle\pm1.60}$ & 0.187 & 0.112 & 0.134 \\
160 & $77.75{\scriptstyle\pm2.63}$ & $82.25{\scriptstyle\pm2.21}$ & $94.00{\scriptstyle\pm1.33}$ & 0.200 & 0.131 & 0.159 \\
168 & $80.50{\scriptstyle\pm2.07}$ & $79.75{\scriptstyle\pm2.23}$ & $93.00{\scriptstyle\pm1.13}$ & 0.190 & 0.134 & 0.165 \\
192 & $86.00{\scriptstyle\pm1.92}$ & $86.00{\scriptstyle\pm1.79}$ & $94.25{\scriptstyle\pm1.01}$ & 0.196 & 0.152 & 0.174 \\
204 & $85.00{\scriptstyle\pm1.94}$ & $92.25{\scriptstyle\pm1.49}$ & $92.75{\scriptstyle\pm1.33}$ & 0.200 & 0.171 & 0.177 \\
240 & $90.00{\scriptstyle\pm1.46}$ & $93.00{\scriptstyle\pm1.55}$ & $95.25{\scriptstyle\pm0.99}$ & 0.208 & 0.188 & 0.188 \\
276 & $92.75{\scriptstyle\pm1.12}$ & $93.50{\scriptstyle\pm1.33}$ & $97.00{\scriptstyle\pm0.83}$ & 0.222 & 0.204 & 0.231 \\
288 & $94.00{\scriptstyle\pm1.02}$ & $92.75{\scriptstyle\pm1.50}$ & $97.75{\scriptstyle\pm0.68}$ & 0.223 & 0.197 & 0.241 \\
\bottomrule
\end{tabular}
\end{table}

\begin{table}[H]
\centering
\caption{Reacher multi-channel same-message: message accuracy (\%)
  ($\sigma{=}1.0$, 8 targets/rollout, $n{=}50$).}
\label{tab:reacher-same}
\begin{tabular}{r rrr rrr}
\toprule
& \multicolumn{3}{c}{Message Accuracy (\%)}
& \multicolumn{3}{c}{Margin} \\
\cmidrule(lr){2-4}\cmidrule(lr){5-7}
$T$ & $M{=}2$ & $M{=}4$ & $M{=}8$
    & $M{=}2$ & $M{=}4$ & $M{=}8$ \\
\midrule
96  & $76.50{\scriptstyle\pm2.71}$ & $59.00{\scriptstyle\pm2.94}$ & $84.75{\scriptstyle\pm1.91}$ & 0.143 & 0.099 & 0.092 \\
128 & $79.00{\scriptstyle\pm2.16}$ & $73.50{\scriptstyle\pm2.41}$ & $93.00{\scriptstyle\pm1.18}$ & 0.201 & 0.124 & 0.138 \\
132 & $77.00{\scriptstyle\pm2.25}$ & $74.50{\scriptstyle\pm2.26}$ & $94.50{\scriptstyle\pm0.95}$ & 0.181 & 0.130 & 0.142 \\
160 & $82.75{\scriptstyle\pm1.87}$ & $89.50{\scriptstyle\pm1.63}$ & $96.25{\scriptstyle\pm0.95}$ & 0.193 & 0.143 & 0.161 \\
168 & $83.00{\scriptstyle\pm1.96}$ & $87.75{\scriptstyle\pm1.85}$ & $97.00{\scriptstyle\pm0.76}$ & 0.188 & 0.148 & 0.166 \\
192 & $86.00{\scriptstyle\pm2.20}$ & $92.75{\scriptstyle\pm1.37}$ & $97.50{\scriptstyle\pm0.71}$ & 0.196 & 0.151 & 0.174 \\
204 & $84.50{\scriptstyle\pm2.31}$ & $92.50{\scriptstyle\pm1.22}$ & $98.25{\scriptstyle\pm0.61}$ & 0.202 & 0.174 & 0.187 \\
240 & $92.75{\scriptstyle\pm1.50}$ & $95.50{\scriptstyle\pm0.98}$ & $98.50{\scriptstyle\pm0.57}$ & 0.205 & 0.185 & 0.194 \\
276 & $94.75{\scriptstyle\pm1.28}$ & $96.50{\scriptstyle\pm0.94}$ & 100.00                       & 0.211 & 0.198 & 0.255 \\
288 & $95.50{\scriptstyle\pm0.98}$ & $96.75{\scriptstyle\pm0.85}$ & $99.75{\scriptstyle\pm0.25}$ & 0.206 & 0.196 & 0.259 \\
\bottomrule
\end{tabular}
\end{table}

\begin{table}[H]
\centering
\caption{Reacher multi-channel split-message with optimal
  allocation ($n{=}50$). Each row uses the minimum $T$
  and $(l_\text{inner}, l_\text{outer})$ allocation satisfying
  the 90\% target predicted from single-channel curves.}
\label{tab:reacher-indep}
\begin{tabular}{r r r r rr rr}
\toprule
$M$ & bits & $(l_\text{inner}, l_\text{outer})$ & $T$ & Msg.\ (\%) & Sym.\ (\%)
    & Dim\,0 (\%) & Dim\,1 (\%) \\
\midrule
8 & 9  & (2,\,1) & 344 & $93.50{\scriptstyle\pm1.20}$ & $97.75{\scriptstyle\pm0.42}$ & $97.25{\scriptstyle\pm0.54}$ & $98.75{\scriptstyle\pm0.54}$ \\
4 & 6  & (2,\,1) & 306 & $92.50{\scriptstyle\pm1.29}$ & $97.25{\scriptstyle\pm0.49}$ & $97.38{\scriptstyle\pm0.59}$ & $97.00{\scriptstyle\pm0.76}$ \\
\bottomrule
\end{tabular}
\end{table}

Splitting pays off when the budget is tight: it beats a single arm by 26 points at $T{=}344$. At $T{=}306$ the single arm is already accurate on its own, and the gap falls to $3.5$ points.

\subsection{Football: Numerical Results}
\label{app:football-results}

\begin{table}[H]
\centering
\caption{Football single-channel (agent~0): goal-decoding accuracy
(\%) and margin vs.\ $T$ and $M$ ($n{=}50$ rollouts,
64~strategies per cell).}
\label{tab:football-single}
\begin{tabular}{r rrr rrr}
\toprule
& \multicolumn{3}{c}{Accuracy (\%)}
& \multicolumn{3}{c}{Margin} \\
\cmidrule(lr){2-4}\cmidrule(lr){5-7}
$T$ & $M{=}4$ & $M{=}8$ & $M{=}64$
    & $M{=}4$ & $M{=}8$ & $M{=}64$ \\
\midrule
96   & 20.2 & 28.5 & 36.3 & 0.07 & 0.06 & 0.03 \\
180  & 24.2 & 35.9 & 35.9 & 0.06 & 0.05 & 0.02 \\
264  & 37.2 & 33.6 & 55.2 & 0.06 & 0.04 & 0.02 \\
348  & 33.2 & 49.8 & 75.4 & 0.05 & 0.04 & 0.04 \\
432  & 48.5 & 56.1 & 91.4 & 0.06 & 0.04 & 0.06 \\
516  & 41.7 & 60.6 & 97.2 & 0.05 & 0.04 & 0.08 \\
\bottomrule
\end{tabular}
\end{table}

\begin{table}[H]
\centering
\caption{Football multi-channel same-message: goal-decoding
accuracy (\%) vs.\ $T$ and $M$ ($n{=}50$ rollouts,
64~strategies per cell).}
\label{tab:football-same}
\begin{tabular}{r rrr rrr}
\toprule
& \multicolumn{3}{c}{Accuracy (\%)}
& \multicolumn{3}{c}{Margin} \\
\cmidrule(lr){2-4}\cmidrule(lr){5-7}
$T$ & $M{=}4$ & $M{=}8$ & $M{=}64$
    & $M{=}4$ & $M{=}8$ & $M{=}64$ \\
\midrule
96   & 41.5 & 55.3 & 68.5  & 0.06 & 0.06 & 0.03 \\
180  & 50.6 & 66.2 & 66.0  & 0.05 & 0.05 & 0.02 \\
264  & 68.2 & 66.0 & 83.7  & 0.06 & 0.04 & 0.03 \\
348  & 65.9 & 82.2 & 96.7  & 0.05 & 0.04 & 0.04 \\
432  & 80.8 & 86.4 & 99.8  & 0.05 & 0.04 & 0.07 \\
516  & 74.2 & 89.6 & 100.0 & 0.04 & 0.04 & 0.09 \\
\bottomrule
\end{tabular}
\end{table}

\begin{table}[H]
\centering
\caption{Football multi-channel split-message ($M{=}4$,
$4^{3}{=}64$ strategies): each agent transmits one quaternary symbol
independently using its own 2D velocity signal ($n{=}50$~rollouts).}
\label{tab:football-indep}
\begin{tabular}{r r r}
\toprule
$T$ & Accuracy (\%) & Margin \\
\midrule
96   & 41.4 & 0.06 \\
180  & 37.2 & 0.04 \\
264  & 55.3 & 0.05 \\
348  & 74.1 & 0.06 \\
432  & 90.7 & 0.08 \\
516  & 97.1 & 0.10 \\
\bottomrule
\end{tabular}
\end{table}

\subsection{RoboMaster: Numerical Results}
\label{app:robomaster-results}

We validate the multi-channel split-message scheme on physical
hardware using four RoboMaster ground robots in a motion-capture
arena. Each robot follows a fixed six-waypoint zigzag trajectory
and transmits its own distinct $M{=}4$ symbol (the index of one of
four top-edge targets it ultimately drives to). The four robots
constitute four independent CGN channels carrying a single 4-symbol
(8-bit) message per rollout. A proportional go-to-goal controller
at $50$\,Hz drives the platform; watermark noise is added on the
2D velocity command ($\sigma{=}0.1$, base speed $0.5$\,m/s).
Detection uses the same coherency-based decoder as the simulation
experiments, run offline on the motion-capture position traces. Each
rollout draws an independent codebook seed and an independent
random permutation mapping robots to targets, so the four reported
decodes per rollout are scored against rollout-specific references.
Results are aggregated over $n{=}20$ rollouts (80 per-robot symbol
decodes per setting); reported errors are standard errors across
rollouts.

\begin{table}[H]
\centering
\caption{RoboMaster multi-channel split-message: per-symbol and
joint-message accuracy (\%) vs.\ $T$ at $M{=}4$,
$\sigma{=}0.1$, 4 robots/rollout ($L{=}4$ symbols,
$(l_1, l_2, l_3, l_4){=}(1,1,1,1)$), $n{=}20$ rollouts.
``Msg.\ Acc.''\ is the joint probability that all four robots'
symbols decode correctly in the same rollout. At $50$\,Hz policy
rate, $T{=}200$\,steps corresponds to $4$\,s of time per symbol.}
\label{tab:robomaster-indep}
\begin{tabular}{r r r r}
\toprule
$T$ & Msg.\ Acc.\ (\%) & Sym.\ Acc.\ (\%) & Margin \\
\midrule
 200 & $65.0{\scriptstyle\pm10.9}$ & $88.8{\scriptstyle\pm4.2}$ & $0.102{\scriptstyle\pm0.007}$ \\
 300 & $95.0{\scriptstyle\pm5.0}$  & $98.8{\scriptstyle\pm1.2}$ & $0.206{\scriptstyle\pm0.010}$ \\
 400 & $95.0{\scriptstyle\pm5.0}$  & $98.8{\scriptstyle\pm1.2}$ & $0.239{\scriptstyle\pm0.011}$ \\
 500 & 100.0                       & 100.0                      & $0.258{\scriptstyle\pm0.011}$ \\
 600 & 100.0                       & 100.0                      & $0.278{\scriptstyle\pm0.011}$ \\
\bottomrule
\end{tabular}
\end{table}

The underlying margin between the correct symbol and its nearest competitor increases monotonically across the sweep, confirming that the decoder's confidence grows continuously with window length.

\subsection{Error-Correcting Codes Analysis}
\label{app:discovery-rs-analysis}
A natural ECC choice for $M$-ary alphabets are Reed--Solomon codes.

We consider applying RS codes at the goal level: since each message
is one of $8$ goal IDs, we apply RS$(n,k)$ over $\mathrm{GF}(8)$,
correcting wrongly-decoded goals rather than individual transmission
symbols.
Table~\ref{tab:rs-overhead} shows the lowest-redundancy code for
each correction capability~$t$ and alphabet size~$M$.

\begin{table}[H]
\centering
\caption{Best-case RS overhead factor for $T$
per data goal, for each correction capability (num.\ errors) $t$ and alphabet
size~$M$.}
\label{tab:rs-overhead}
\begin{tabular}{c c c c c}
\toprule
$t$ & Best code & $M{=}8$ & $M{=}4$ & $M{=}2$ \\
\midrule
1 & RS$(7,5)$ & $1.4\times$ & $2.8\times$ & $4.2\times$ \\
2 & RS$(7,3)$ & $2.3\times$ & $4.7\times$ & $7.0\times$ \\
3 & RS$(7,1)$ & $7.0\times$ & $14\times$  & $21\times$  \\
\bottomrule
\end{tabular}
\end{table}

Each multiplier increases $T$ by the same factor. Whether
RS coding pays off depends on the shape of the per-environment accuracy
curve: spending the same total budget on a longer uncoded
$T$ often matches or beats RS, particularly where the
same-message curve is still climbing and a $1.4$--$7\times$ increase in
$T$ pushes per-goal accuracy close to~1. RS only tends to help when the uncoded curve has plateaued below~1, so
that further increasing $T$ no longer reliably eliminates
goal errors and the spare budget is better spent on parity goals.

\subsection{Reward Definitions}
\label{app:reward-defs}

Exact metrics behind Table~\ref{tab:reward-impact} (Sec.~\ref{app:reward-impact}).
In VMAS Discovery, we use final Euclidean distance
$\|p_{\mathrm{agent}} - p_{\mathrm{goal}}\|$. Lunar Lander uses Box2D's standard reward
(distance shaping to the landing pad, $-0.3$ per main-engine step, $-0.03$
per side-engine step, $+100$ on landing, $-100$ on crash), summed over the
episode. Reacher uses MuJoCo's dense reward
$r_t = -\|p_{\mathrm{fingertip}} - p_{\mathrm{target}}\| - 0.1\|a_t\|^2$, summed over the episode. In VMAS Football, we report the mean distance from the ball to the attacking
goal at $(+1.51,\,0)$. Energy and actuator wear were not measured.

\subsection{Full Attack Sweeps}
\label{app:attacks-full}

Tables~\ref{tab:adv-additive} and~\ref{tab:adv-bandstop} give the full sweeps
summarized in Table~\ref{tab:attacks} (Sec.~\ref{app:adversarial}); setup as
there ($M{=}8$, $T{=}120$, $\sigma_{\mathrm{wm}}{=}0.03$, $n{=}50$ i.i.d.\ rollouts per cell).

\noindent\textbf{Additive noise.}
Message accuracy is intact for $\sigma_{\mathrm{adv}} \le \sigma_{\mathrm{wm}}$,
partial at $\sigma_{\mathrm{adv}}{\approx}1.7\sigma_{\mathrm{wm}}$, and settles a few
points above the $1/M$ chance rate from $\sigma_{\mathrm{adv}}{\approx}5\sigma_{\mathrm{wm}}$
onward. The low task cost on Discovery does not carry over to tasks with a terminal
success criterion: on Lunar Lander ($M{=}8$, $T{=}288$, $\sigma_{\mathrm{wm}}{=}0.5$), at
$0.83\sigma_{\mathrm{wm}}$ message accuracy is still $100\%$ while the landing rate has
collapsed from $79.2\%$ to $6.8\%$.

\begin{table}[H]
\centering
\caption{Additive-noise attack on Discovery (full sweep).}
\label{tab:adv-additive}
\begin{tabular}{r r r r}
\toprule
$\sigma_{\mathrm{adv}}/\sigma_{\mathrm{wm}}$ & Msg.\ acc.\ (\%) & $\Delta$cmd (\%) & $\Delta$dist (\%)\\
\midrule
0.00 & 100.0 &   0.0 & 0.0 \\
0.33 & 100.0 &   7.2 & 0.0 \\
0.83 &  98.5 &  18.1 & 0.0 \\
1.67 &  67.0 &  36.2 & 0.1 \\
3.33 &  24.5 &  72.4 & 0.3 \\
5.00 &  16.5 & 108.5 & 0.5 \\
6.67 &  15.5 & 144.3 & 0.8 \\
\bottomrule
\end{tabular}
\end{table}

\noindent\textbf{Band-stop filtering.}
Full sweep over Butterworth filter order, with the detector margin.

\begin{table}[H]
\centering
\caption{Band-stop attack on Discovery with $B$ overlapping the
policy's dominant frequencies (full sweep). $\Delta$cmd is
$\mathrm{RMS}(a_{\mathrm{post}}-a_{\mathrm{pre}})/\mathrm{RMS}(a_{\mathrm{pre}})$;
$\Delta$dist is the increase in mean agent--goal distance vs.\
the order-0 baseline.}
\label{tab:adv-bandstop}
\begin{tabular}{r c c c c}
\toprule
Filter order & Msg.\ Acc.\ (\%) & Margin & $\Delta$cmd (\%) & $\Delta$dist (\%) \\
\midrule
0  & 100.0 & 0.128 &  0.0 &  0.0 \\
2  & 22.5  & 0.035 & 36.2 &  4.3 \\
4  & 14.0  & 0.033 & 46.9 &  8.1 \\
6  & 12.3  & 0.031 & 53.6 & 12.2 \\
8  & 14.0  & 0.031 & 59.9 & 16.7 \\
10 & 12.0  & 0.030 & 66.5 & 21.5 \\
\bottomrule
\end{tabular}
\end{table}

\noindent\textbf{Structured jamming.}
Because the watermark is colored rather than white, its samples are correlated
across time, so a learned jammer $J$ conditioned on the current observation and
recent motion history could in principle become correlated with the watermark
$W$---and partially predict it---from observed motion alone, without recovering
the key. The covariance term in
$\mathrm{Var}(W + J) = \mathrm{Var}(W) + \mathrm{Var}(J) + 2\,\mathrm{Cov}(W, J)$
therefore need not vanish. We do not evaluate this attack.

\noindent\textbf{Distillation.}
Behavior-cloning the watermarked policy onto a fresh surrogate could
in principle strip the watermark as a side-effect of imperfect
approximation: MTM's per-symbol signals live in the temporal
correlation structure of the policy's exploration noise rather than its marginal action distribution, so a surrogate that samples fresh exploration noise at each step would not preserve the keyed CGN that carries messages. 
Two practical constraints limit this attack: (i) the adversary might not realistically be able to collect enough data to clone the
policy well: the original was trained on huge, private datasets,
whereas the adversary can only sample its behavior by deploying it on
a few robots or making API queries, which could be costly and have limited usage, and (ii) the performance degradation inherent in imperfect distillation~\cite{cho2019efficacy}, which reduces the value of the stolen IP. We do not experiment with this attack.

\end{document}